\pdfoutput=1

\documentclass[journal]{IEEEtai}

\usepackage{cite}
\usepackage{amsmath,amssymb,amsfonts}
\usepackage{amsthm}
\usepackage{graphicx}
\usepackage{textcomp}
\usepackage{xcolor}
\usepackage{booktabs}
\usepackage{multirow}
\usepackage{array}
\usepackage{tabularx}
\usepackage{longtable}
\usepackage{threeparttable}
\usepackage{url}
\usepackage[colorlinks,urlcolor=blue,linkcolor=blue,citecolor=blue]{hyperref}

\graphicspath{{Fig/}}

\newcommand{\figalttext}[2][]{}

\begin{document}

\title{Structure-Guided Adaptive Propagation for Protein-Protein Interaction Site Prediction}

\author{Enqiang~Zhu, Yizi~Liu, Yilong~Luo, Yao~Chen, Yu~Zhang, and Baoshan~Ma%
\thanks{Enqiang Zhu and Yizi Liu are with the Institute of Computing Science and Technology, Guangzhou University, Guangzhou 510006, Guangdong, China.}%
\thanks{Yilong Luo and Yu Zhang are with the School of Computer Science, Peking University, Beijing 100871, China.}%
\thanks{Yao Chen is with the Information Science \& Technology Department, Beijing Capital International Airport Co., Ltd., Beijing 101317, China.}%
\thanks{Baoshan Ma is with the School of Information Science and Technology, Dalian Maritime University, Dalian 116026, China.}%
\thanks{Corresponding authors: Yu Zhang (yuzhang.cs@stu.pku.edu.cn) and Baoshan Ma (mabaoshan@dlmu.edu.cn).}%
}

\markboth{arXiv preprint, 2026}%
{Zhu \MakeLowercase{\textit{et al.}}: Structure-Guided Adaptive Propagation for Protein-Protein Interaction Site Prediction}

\maketitle

\begin{abstract}
 Accurate prediction of protein-protein interaction sites (PPIS) is essential for understanding cellular processes, disease mechanisms, and therapeutic target discovery. Graph-based deep learning has advanced PPIS prediction by incorporating residue-level structural context. However, most graph-based models still rely on fixed propagation schemes that treat all residues similarly, despite the structural and functional heterogeneity of protein interfaces. Such propagation may limit the ability to adapt information diffusion to local geometric environments, making it difficult to distinguish true interaction sites from structurally similar non-interacting neighbors. We present SGAP-PPIS, a structure-guided adaptive propagation model for PPIS prediction. Rather than using a fixed propagation mechanism, SGAP-PPIS leverages multi-scale geometric states from an equivariant graph neural network to generate residue-wise propagation coefficients. This design allows each residue to adaptively balance local feature preservation and neighborhood diffusion according to its geometric microenvironment. Experimental results show that SGAP-PPIS achieves competitive performance among the state-of-the-art methods on Test\_60. Ablation studies show that geometry-conditioned adaptive propagation, scale-aligned geometric guidance, and multi-step propagation-state representation jointly drive these improvements.
\end{abstract}


\section{Introduction}
Proteins are central to the execution of life's activities and play vital roles in numerous key processes within living organisms, including signal transduction~\cite{pawson2000protein}, nutrient transport~\cite{hubel2019protein}, and cellular metabolism~\cite{wirtz2001cysteine}. 
The protein-protein interaction site (PPIS) is crucial for regulating these biological processes~\cite{li2010mouse}. Abnormalities in PPIS are closely linked to the development of various diseases, such as cancer and autoimmune disorders~\cite{zhang2018review}. Therefore, investigating PPIS is of great importance in uncovering cellular processes, enhancing our understanding of disease mechanisms, and facilitating the design and discovery of new drugs~\cite{zhang2025edg}. While traditional experimental techniques, such as X-ray crystallography and yeast two-hybrid screening, have generated valuable experimental data, their high costs and complex workflows limit their widespread application~\cite{hamp2015more,shoemaker2007deciphering}. These limitations have motivated the development of computational methods for PPIS prediction as an important complement to experimental approaches.

Machine learning has long been used for predicting PPIS. Early methodologies in this domain include Naive Bayes classifiers~\cite{murakami2010applying}, Random Forests~\cite{northey2018intpred}, and Logistic Regression \cite{zhang2019scriber}, among others. These methods rely on feature engineering to represent proteins using descriptors such as raw sequences, position-specific scoring matrices (PSSMs), and DSSP-derived secondary-structure features. Early studies showed that such traditional machine-learning approaches could achieve meaningful results due to their interpretability and their ability to incorporate domain knowledge. Their reliance on handcrafted features limits their capacity to capture complex residue-level interactions.

In recent years, deep learning methods have been increasingly applied to improve the accuracy of PPIS prediction. Existing deep learning approaches can be broadly categorized into two classes: sequence-based methods and structure-based methods.

Sequence-based deep learning methods have been widely used for PPIS prediction.
For example, ConvsPPIS~\cite{zhu2020convsppis} is a Convolutional Neural Network (CNN)-based sequence method that constructs feature graphs along protein sequences and uses CNNs to capture local residue context and positional patterns. 
DELPHI~\cite{li2021delphi} is a hybrid sequence-based method that combines CNN and RNN to extract complementary information from multiple sequence-derived feature groups for PPIS prediction.

Sequence-based methods are inherently limited in their ability to explicitly model the three-dimensional spatial relationships between residues. Structure-based methods address this limitation by incorporating protein structural information into PPIS prediction. 
For example, GraphPPIS~\cite{yuan2022structure} represents a protein as an undirected protein graph and treats PPIS prediction as a node classification task by integrating evolutionary and structural features with pairwise residue-distance information. ASCE-PPIS~\cite{shen2025asce} enhances structural modeling through equivariant message passing, structure-aware pooling, graph collapse, and ensemble learning. MGMA-PPIS~\cite{han2025mgma} constructs multi-view graph representations by combining equivariant GNN-based global features with multi-scale local features. ComGAT-PPIS~\cite{zhang2025comgat} introduces community-enhanced graph attention to integrate residue-level and community-level structural information. 
Despite these advances, most existing models improve PPIS prediction primarily through stronger feature extraction, hierarchical abstraction, or post-fusion strategies, while the graph-propagation mechanism itself remains largely fixed. However, protein interface residues are not structurally homogeneous. Previous studies have shown that protein interfaces can be partitioned into support, core, and rim regions with distinct structural and functional properties~\cite{guharoy2005conservation,levy2010simple,david2015contribution,laine2015local}. This interface heterogeneity suggests that different residues may require different balances between feature preservation and neighborhood diffusion during graph propagation.

To address these limitations, we introduce SGAP-PPIS, a structure-guided adaptive propagation model designed for PPIS prediction. The fundamental concept of SGAP-PPIS is that different residues should exhibit distinct propagation behaviors during graph learning. Rather than applying a uniform approach, the model adaptively determines the balance between feature preservation and neighborhood diffusion based on the local geometric environment of each residue. To achieve this, SGAP-PPIS first utilizes an equivariant geometric encoder to extract multi-scale geometric states from the protein graph. These geometric states are then employed to generate residue-specific propagation coefficients for adaptive APPNP-based information diffusion~\cite{gasteiger2019predict}. Furthermore, the model retains intermediate propagation states and combines them with the geometric summary to create more discriminative residue representations for final predictions. We conducted experiments to assess the performance of SGAP-PPIS. The experimental results show that SGAP-PPIS outperforms existing methods across these datasets, achieving the highest values on all evaluated metrics. For instance, on the primary dataset Test\_60, SGAP-PPIS attains an MCC of 0.555, an AUROC of 0.894, and an AUPRC of 0.670.

\section{Materials and methods}\label{sec2}
This section outlines the benchmark datasets, protein graph construction, and the methodological design of SGAP-PPIS. 

\subsection{Datasets}\label{subsec:Datasets}
The datasets used in this study follow the benchmark setting adopted in GraphPPIS~\cite{yuan2022structure}, comprising the training set Train\_335-1 and three independent test sets: Test\_60, Test\_315-28, and UBtest\_31-6. Test\_60 was used as the primary benchmark for performance comparison, whereas Test\_315-28 and UBtest\_31-6 were used to evaluate generalization ability. The statistics for the processed datasets are summarized in Table~\ref{tab:datasets}, and further details on dataset construction and preprocessing are provided in the Supplementary Material.
 
\begin{table}[ht]
\centering
\caption{Dataset statistics}
\resizebox{\linewidth}{!}{
\begin{tabular}{lccc}
\hline
Dataset   & Interacting & Non-interacting & Interaction ratio(\%) \\
\hline
Train\_335-1  & 10,336 & 55,872 & 15.61 \\
Test\_60    & 2,075  & 11,069 & 15.79 \\
Test\_315-28  & 8,566  & 51,810 & 14.19 \\
UBtest\_31-6   & 711    & 5,206  & 12.02 \\
\hline
\end{tabular}
}
\label{tab:datasets}
\end{table}

 \begin{figure*}[htb]
    \centering
    \includegraphics[width=1.0\textwidth]{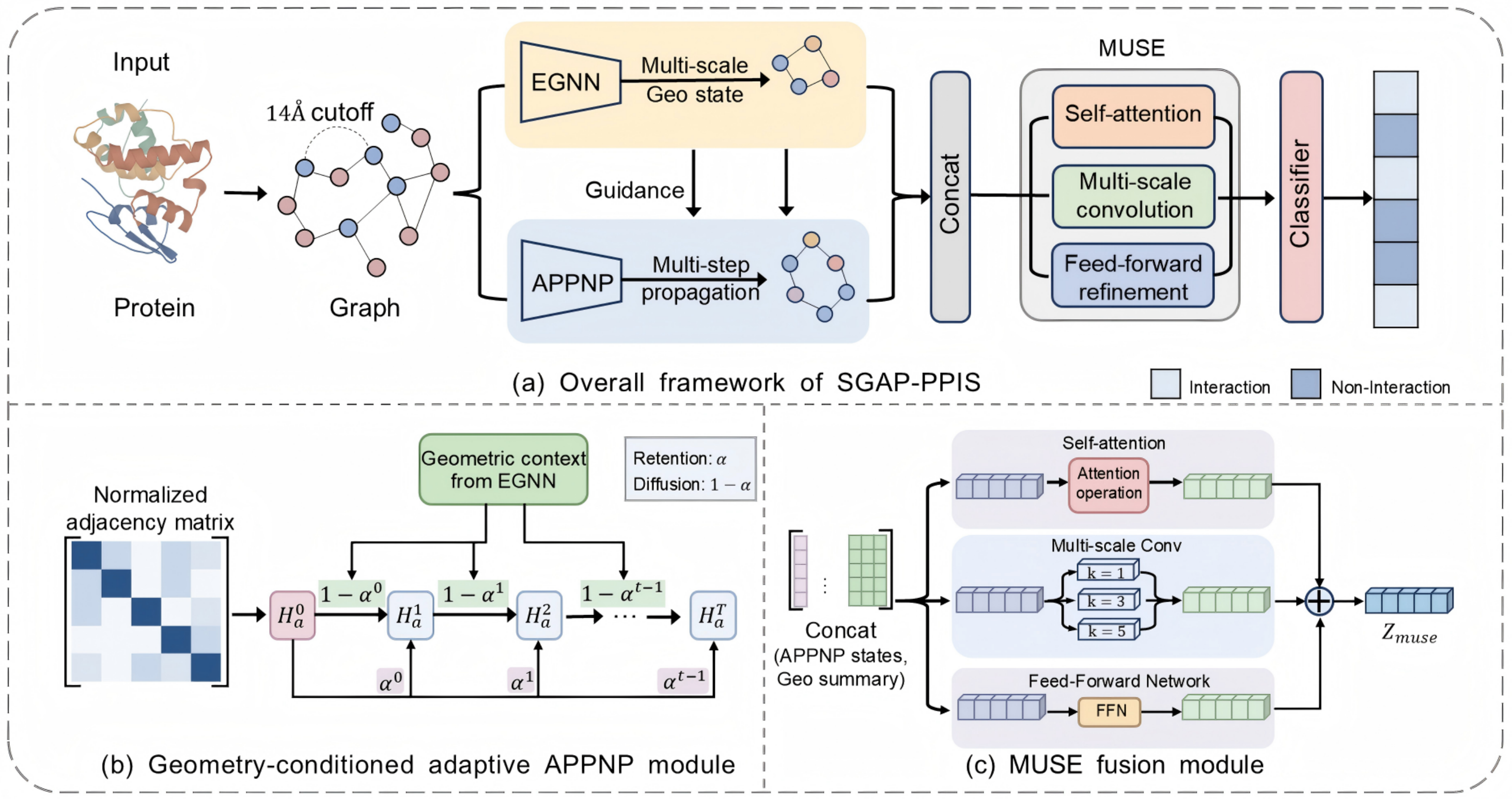}
    \caption{Overall architecture of SGAP-PPIS. (a) Dual-branch model combining EGNN-based geometric encoding and geometry-conditioned APPNP propagation. (b) Adaptive propagation with residue-wise coefficients and multi-step state retention. (c) Multi-scale attention fusion via self-attention, convolution, and feed-forward layers.}
    \label{fig:1}
\end{figure*}

\subsection{Model architecture}\label{subsec:Model architecture}
In this study, PPIS prediction is formulated as a node-level binary classification problem on protein graph. The SGAP-PPIS model is illustrated in Figure~\ref{fig:1}. 
Each protein is represented as an undirected graph \(G = (V, E) \), where \( V \) denotes the set of residues and \( E \) denotes the set of edges connecting pairs of residues. The relative coordinates of nodes in 3D space, after centralization, are represented as the pseudo-coordinate \( P \in \mathbb{R}^{N \times 3} \). An edge is constructed between two residues if the Euclidean distance \(d_{ij}\) between their side-chain centroids is below a predefined threshold. The node feature matrix, which encapsulates all node features in the graph, is represented as \( X \in \mathbb{R}^{N \times 62} \), where each node is characterized by a feature vector \( x_i \in \mathbb{R}^{62} \). Similarly, the feature matrix for all edge features in the graph is represented as \( Y \in \mathbb{R}^{M \times 2} \), where each edge is associated with a feature vector \( y_{ij} \in \mathbb{R}^2 \). Here, \( N \) and \( M \) correspond to the number of residues and edges in the protein, respectively.

SGAP-PPIS processes a protein graph through five sequential stages. It begins with graph construction and feature initialization. A multi-scale geometric encoding is then performed via an EGNN branch, followed by geometry-conditioned adaptive propagation through an APPNP branch. The resulting cross-view features are fused in a MUSE module, after which a final residue-level classification head produces class logits that can be converted into interaction probabilities.

SGAP-PPIS uses multi-scale geometric context not only for feature fusion but also to regulate residue-level propagation.
The EGNN branch extracts geometric states from protein graph at each layer and aggregates them into a multi-scale geometric summary.
The APPNP branch performs multi-step propagation, where residue-wise propagation coefficients are generated from the geometric context produced by the EGNN branch.

To preserve information across different propagation steps, SGAP-PPIS retains the intermediate APPNP states and concatenates them to form the propagation representation. This representation is then combined with the multi-scale geometric summary and further fused by the MUSE module, which integrates self-attention, multi-scale convolution, and a position-wise feed-forward network. Finally, the fused representation is fed into the classification head to predict PPIS.

\subsection{Protein representation}\label{subsec:Protein representation}

In SGAP-PPIS, each protein is represented as an undirected protein graph \(G = (V, E)\), where residues are modeled as nodes and spatially neighboring residue pairs are connected by edges. 

1) Node Features: Following previous PPIS prediction studies~\cite{zeng2024ghgpr,zhou2023agat}, we adopt a 62-dimensional node feature set in this work, which includes evolutionary sequence features: Position-Specific Scoring Matrix (PSSM)~\cite{altschul1997gapped} and Hidden Markov Model (HMM) matrix~\cite{remmert2012hhblits}; structural features: Protein Secondary Structure features (DSSP)~\cite{kabsch1983dictionary}, residue pseudo-position embedding (PPE), and Atomic features (AFs) of the residues.

For a protein with \(N\) residues, the node feature matrix is denoted as \(X \in \mathbb{R}^{N \times 62}\).
{
\abovedisplayskip=3pt plus 0pt minus 0pt
\belowdisplayskip=3pt plus 0pt minus 0pt
\abovedisplayshortskip=1pt plus 1pt minus 1pt
\belowdisplayshortskip=1pt plus 1pt minus 1pt
\begin{align}
X &= [X_{\text{PSSM}}, X_{\text{HMM}}, X_{\text{DSSP}}, X_{\text{AF}}, X_{\text{PPE}}] \label{eq2} 
\end{align}
}
More detailed descriptions of these handcrafted features are provided in the Supplementary Material.

2) Graph Connectivity: The pairwise distance matrix is computed using the Euclidean distance between the side-chain centroids of two residues. Following previous studies, an edge is created when the distance is less than \( \tau\), where the cutoff threshold is set to \(\tau = 14\,\text{\AA}\). The adjacency matrix \(A\) is then defined by \(A_{ij}=1\) if residues \(i\) and \(j\) are connected, and \(A_{ij}=0\) otherwise. This construction connects spatially close residues and preserves structural contacts relevant to PPIS prediction.

3) Edge Features: Each edge is associated with a 2-dimensional geometric feature vector \( y_{ij} \in \mathbb{R}^{2} \), consisting of the normalized Euclidean distance between residues \(i\) and \(j\) and the normalized cosine similarity between their pseudo-position vectors.

\subsection{Equivariant graph neural network}\label{subsec:Equivariant graph neural network}
Equivariant Graph Neural Networks (EGNNs) are suitable for protein structure modeling because they update node representations and coordinates while preserving equivariance to Euclidean transformations.

In SGAP-PPIS, the EGNN~\cite{satorras2021n} branch serves as an equivariant geometric encoder for extracting multi-scale geometric states from protein graph. The raw node feature matrix \(X\) is first projected into a geometric hidden space:
{
\abovedisplayskip=3pt plus 0pt minus 0pt
\belowdisplayskip=3pt plus 0pt minus 0pt
\abovedisplayshortskip=1pt plus 1pt minus 1pt
\belowdisplayshortskip=1pt plus 1pt minus 1pt
\begin{align}
H_g^{(0)} &= \mathrm{ReLU}(XW_g + b_g)
\label{eq3}
\end{align}
} Here, \(H_g^{(0)} \in \mathbb{R}^{N \times d_g}\) denotes the initial hidden representation of the EGNN branch, where \(N\) is the number of residues and \(d_g\) is the hidden dimension in the EGNN branch.

Starting from \((H_g^{(0)}, P^{(0)})\), this model stacks \(K\) EGNN layers to update node representations and pseudo-coordinates through equivariant message passing. For the \(k\)-th layer, the update process is defined as:

{
\abovedisplayskip=3pt plus 0pt minus 0pt
\belowdisplayskip=3pt plus 0pt minus 0pt
\abovedisplayshortskip=1pt plus 1pt minus 1pt
\belowdisplayshortskip=1pt plus 1pt minus 1pt
\begin{align}
m_{ij}^{(k)} &= \phi_e\bigl(h_i^{(k)}, h_j^{(k)}, \|p_i^{(k)} - p_j^{(k)}\|^2, e_{ij}\bigr)
\label{eq4}
\end{align}
}
{
\abovedisplayskip=3pt plus 0pt minus 0pt
\belowdisplayskip=3pt plus 0pt minus 0pt
\abovedisplayshortskip=1pt plus 1pt minus 1pt
\belowdisplayshortskip=1pt plus 1pt minus 1pt
\begin{align}
\tilde{m}_{ij}^{(k)} &= \sigma\bigl(\phi_{\mathrm{att}}(m_{ij}^{(k)})\bigr) \odot m_{ij}^{(k)}
\label{eq5}
\vspace{-5pt}
\end{align}
}
{
\abovedisplayskip=3pt plus 0pt minus 0pt
\belowdisplayskip=3pt plus 0pt minus 0pt
\abovedisplayshortskip=1pt plus 1pt minus 1pt
\belowdisplayshortskip=1pt plus 1pt minus 1pt
\begin{align}
p_i^{(k+1)} &= p_i^{(k)} + \mathrm{Mean}_{j \in \mathcal{N}(i)}
\left[
\bigl(p_i^{(k)} - p_j^{(k)}\bigr)\phi_p\bigl(\tilde{m}_{ij}^{(k)}\bigr)
\right]
\label{eq6}
\end{align}
}
{
\abovedisplayskip=3pt plus 0pt minus 0pt
\belowdisplayskip=3pt plus 0pt minus 0pt
\abovedisplayshortskip=1pt plus 1pt minus 1pt
\belowdisplayshortskip=1pt plus 1pt minus 1pt
\begin{align}
m_i^{(k)} &= \sum_{j \in \mathcal{N}(i)} \tilde{m}_{ij}^{(k)}
\label{eq7}
\end{align}
}
{
\abovedisplayskip=3pt plus 0pt minus 0pt
\belowdisplayskip=3pt plus 0pt minus 0pt
\abovedisplayshortskip=1pt plus 1pt minus 1pt
\belowdisplayshortskip=1pt plus 1pt minus 1pt
\begin{align}
\bar{h}_i^{(k+1)} &= \phi_h\bigl([h_i^{(k)}, m_i^{(k)}]\bigr), 
\quad h_i^{(k+1)} = \bar{h}_i^{(k+1)} + h_i^{(k)}
\label{eq8}
\end{align}
}
Here, \(\phi_e\), \(\phi_{\mathrm{att}}\), \(\phi_p\), and \(\phi_h\) denote the edge message function, attention gate, coordinate update function, and node feature update function, respectively. The $\sigma(\cdot)$ denotes the sigmoid function for edge-wise message gating.

Each EGNN layer output is mapped to the APPNP hidden dimension by an independent linear layer, followed by ReLU activation, yielding $S^{(k)} \in \mathbb{R}^{N \times d_a}$.

Instead of relying on the output of a single EGNN layer, SGAP-PPIS preserves the geometric states from all EGNN layers and aggregates them through learnable weights:
{
\abovedisplayskip=3pt plus 0pt minus 0pt
\belowdisplayskip=3pt plus 0pt minus 0pt
\abovedisplayshortskip=1pt plus 1pt minus 1pt
\belowdisplayshortskip=1pt plus 1pt minus 1pt
\begin{align}
J &= \sum_{k=1}^{K} \omega_k S^{(k)}, 
\quad J \in \mathbb{R}^{N \times d_a}
\label{eq10}
\end{align}
}
Here, \(S^{(k)}\) denotes the geometric state extracted by the \(k\)-th EGNN layer, and \(J\) represents the multi-scale geometric summary. 
This integrates geometric information across different depths, thereby capturing both local residue-level conformations and higher-order structural organization for subsequent geometry-conditioned adaptive propagation.

\subsection{Approximate personalized propagation of neural predictions}\label{subsec:Approximate personalized propagation of neural predictions}

Graph Convolutional Networks (GCNs) update node representations by aggregating information from local neighborhoods. However, stacking multiple GCN~\cite{kipf2017semi} layers to capture long-range dependencies can lead to over-smoothing, in which node representations become less distinguishable~\cite{li2018deeper}. This limitation motivates propagation-based methods such as APPNP~\cite{gasteiger2019predict}, which decouple feature transformation from graph propagation. In practice, APPNP approximates personalized PageRank through power iteration, enabling efficient multi-step propagation without explicitly computing the full PageRank matrix. To make topological diffusion aware of protein three-dimensional geometry, SGAP-PPIS introduces a geometry-conditioned adaptive propagation mechanism that uses multi-scale geometric states from the EGNN branch to regulate APPNP propagation. 

APPNP alleviates over-smoothing by introducing a teleportation mechanism inspired by personalized PageRank~\cite{page1998pagerank}. During multi-step propagation, each node aggregates neighborhood information while retaining a certain proportion of its root representation. This balance between neighborhood diffusion and feature preservation is particularly important for PPIS prediction, where residues need to incorporate broader structural context without losing discriminative local characteristics.

Given a protein graph \( G = (V, E) \), its original adjacency matrix is \( A \). To ensure the stability of the information propagation, this study first performs symmetric normalization of the adjacency matrix:
{
\abovedisplayskip=3pt plus 0pt minus 0pt
\belowdisplayskip=3pt plus 0pt minus 0pt
\abovedisplayshortskip=1pt plus 1pt minus 1pt
\belowdisplayshortskip=1pt plus 1pt minus 1pt
\begin{align}
\hat{A} &= D^{-1/2} (A+I_N) D^{-1/2}  \label{eq11} 
\end{align}
}
As shown in Equation~\ref{eq11}, \( D \) is the degree matrix of the graph \( D_{ii} = \sum_j (A+I_N)_{ij} \).

The initial node state \( H_a^{(0)} \) of the APPNP module is obtained by applying a linear transformation and activation function to the original node features \( X \):
{
\abovedisplayskip=3pt plus 0pt minus 0pt
\belowdisplayskip=3pt plus 0pt minus 0pt
\abovedisplayshortskip=1pt plus 1pt minus 1pt
\belowdisplayshortskip=1pt plus 1pt minus 1pt
\begin{align}
H_a^{(0)} &= \text{ReLU}(XW_0 + b_0) \label{eq12}
\end{align}
}
As shown in Equation~\ref{eq12}, \( W_0 \) and \( b_0 \) are the learnable parameters of the fully connected layer. \( H_a^{(0)} \) serves as the root information throughout the propagation process.
Before the \(t\)-th APPNP propagation step, SGAP-PPIS uses the EGNN branch to provide scale-aligned geometric guidance. Since the number of APPNP propagation steps and EGNN layers may differ, a layer-wise index mapping is introduced to align each propagation step with a corresponding geometric scale. Specifically, the scale index \( idx \) of the \( t \)-th layer of APPNP corresponding to EGNN is calculated as follows:
{
\abovedisplayskip=3pt plus 0pt minus 0pt
\belowdisplayskip=3pt plus 0pt minus 0pt
\abovedisplayshortskip=1pt plus 1pt minus 1pt
\belowdisplayshortskip=1pt plus 1pt minus 1pt
\begin{align}
\mathrm{idx}(t)=1+\mathrm{round}\!\left(
(t-1)\times\frac{L_{\mathrm{EGNN}}-1}{L_{\mathrm{APPNP}}-1}
\right)  
\end{align}
}
where \(t=1,2,\dots,L_{\mathrm{APPNP}}\) and \(\mathrm{idx}(t)\in\{1,2,\dots,L_{\mathrm{EGNN}}\}\). In this work, $L_{\mathrm{EGNN}}$ denotes the total number of EGNN layers, and $L_{\mathrm{APPNP}}$ denotes the total number of APPNP propagation layers. And \(\mathrm{round}(\cdot)\) denotes rounding to the nearest integer to map each APPNP layer to the closest EGNN scale. After completing the layer alignment, the model computes the geometric context \( C^{(t)} \). This variable is the mean of the EGNN state at the aligned scale \( S_{\text{EGNN}}^{(\text{idx})} \) and the multi-scale geometric summary \( J \):
{
\abovedisplayskip=3pt plus 0pt minus 0pt
\belowdisplayskip=3pt plus 0pt minus 0pt
\abovedisplayshortskip=1pt plus 1pt minus 1pt
\belowdisplayshortskip=1pt plus 1pt minus 1pt
\begin{align}
C^{(t)} &= \frac{1}{2} \left( S_{\text{EGNN}}^{(\text{$idx$})} + J\right)  
\end{align}
}

Traditional APPNP sets the residue-wise propagation coefficient as a hyperparameter. In this model, the proportion of each node that retains the root information during information diffusion should depend on its geometric environment in 3D space. An adaptive residue-wise propagation coefficient $\tilde{\alpha}^{(t)} \in \mathbb{R}^{N \times 1}$ is generated for each node in the graph using the aforementioned geometric context \( C^{(t)} \):
{
\abovedisplayskip=3pt plus 0pt minus 0pt
\belowdisplayskip=3pt plus 0pt minus 0pt
\abovedisplayshortskip=1pt plus 1pt minus 1pt
\belowdisplayshortskip=1pt plus 1pt minus 1pt
\begin{align}
\tilde{\alpha}^{(t)} &= \text{Sigmoid}(W_{\alpha}C^{(t)} + b_{\alpha}) 
\end{align}
}
To avoid extreme propagation behavior, $\tilde{\alpha}$ is clipped into a predefined range: (\( \alpha_{\text{min}} = 0.05 \), \( \alpha_{\text{max}} = 0.95 \)) through the Clamp operation:
{
\abovedisplayskip=3pt plus 0pt minus 0pt
\belowdisplayskip=3pt plus 0pt minus 0pt
\abovedisplayshortskip=1pt plus 1pt minus 1pt
\belowdisplayshortskip=1pt plus 1pt minus 1pt
\begin{align}
\alpha^{(t)} &= \text{Clamp}(\tilde{\alpha}^{(t)}, \alpha_{\text{min}}, \alpha_{\text{max}}) 
\end{align}
Using $\alpha^{(t)}$, the update rule for the \( (t+1) \)-$th$ step propagation in APPNP is defined as:
\begin{align}
H_a^{(t+1)} &= (1 - {\alpha}^{(t)}) \odot (\hat{A} H_a^{(t)}) + {\alpha}^{(t)} \odot H_a^{(0)}  
\end{align}
}
The first term represents neighborhood diffusion over the normalized graph, whereas the second term retains residue-specific root information through the teleportation term. The operator \(\odot\) denotes element-wise multiplication. Here, ${\alpha}^{(t)} \in \mathbb{R}^{N \times 1}$ is broadcast along the feature dimension.

After iterating through \( T \) propagation layers, the model concatenates the hidden layer states from each propagation step along the feature dimension to form the final APPNP topological feature representation:
{
\abovedisplayskip=3pt plus 0pt minus 0pt
\belowdisplayskip=3pt plus 0pt minus 0pt
\abovedisplayshortskip=1pt plus 1pt minus 1pt
\belowdisplayshortskip=1pt plus 1pt minus 1pt
\begin{align}
H_a^{\text{out}} &= \text{$Concat$}(H_a^{(1)}, H_a^{(2)}, \dots, H_a^{(T)} )
\end{align}
}
This multi-layer concatenated feature is then passed into the MUSE module for deep feature fusion with the multi-scale geometric summary from the EGNN branch.

\subsection{Multi-scale attention network}\label{subsec5}

In SGAP-PPIS, the MUSE module is introduced to further fuse the integrated topological and geometric representation. These two representations are concatenated along the feature dimension to form the fused representation:
{
\abovedisplayskip=3pt plus 0pt minus 0pt
\belowdisplayskip=3pt plus 0pt minus 0pt
\abovedisplayshortskip=1pt plus 1pt minus 1pt
\belowdisplayshortskip=1pt plus 1pt minus 1pt
\begin{align}
Z &= [H_a^{\mathrm{out}} \parallel J] \in \mathbb{R}^{N \times (T+1)d_a}
\label{eq18}
\end{align}
}
where \(H_a^{\mathrm{out}}\) encodes topological diffusion information from multiple APPNP propagation steps, and \(J\) denotes the multi-scale geometric summary from the EGNN branch.

The MUSE module captures local and global features in parallel to enhance PPIS recognition \cite{zhao2019muse}. MUSE fuses \(Z\) by summing three parallel branches: a self-attention branch for global dependency modeling, a dynamically weighted multi-scale convolution branch for local pattern extraction, and a position-wise feed-forward network for residue-level feature transformation.

First, a single-head self-attention branch captures long-range dependencies among residues. Given the fused representation \(Z\), it is projected into query, key, and value matrices, and the attention output is computed as:
{
\abovedisplayskip=3pt plus 0pt minus 0pt
\belowdisplayskip=3pt plus 0pt minus 0pt
\abovedisplayshortskip=1pt plus 1pt minus 1pt
\belowdisplayshortskip=1pt plus 1pt minus 1pt
\begin{align}
\mathrm{Attention}(Z) &= \mathrm{softmax}(ZW_Q, ZW_K, ZW_V)W_O
\label{eq19}
\end{align}
}
where \(W_Q\), \(W_K\), \(W_V\), and \(W_O\) are learnable projection matrices, and $\mathrm{softmax}(\cdot)$  denotes the self-attention operation. In the current implementation, residues are arranged in their original sequence order, and each protein is processed individually.

Second, the multi-scale convolution branch extracts local residue patterns using depthwise separable convolutions with different kernel sizes~\cite{wu2019pay}. In this work, 1D depthwise separable convolutions with kernel sizes of 1, 3, and 5 are applied along the residue dimension to capture local contextual patterns at different scales. For a kernel size \(q_i\), the convolutional output is defined as:
{
\abovedisplayskip=3pt plus 0pt minus 0pt
\belowdisplayskip=3pt plus 0pt minus 0pt
\abovedisplayshortskip=1pt plus 1pt minus 1pt
\belowdisplayshortskip=1pt plus 1pt minus 1pt
\begin{align}
\mathrm{Conv}_{q_i}(Z) &= \mathrm{DepthConv}_{q_i}(ZW_V)W_{\mathrm{out}}
\label{eq20}
\end{align}
}
where \(ZW_V\) is the same value representation used in the self-attention branch, and \(W_{\mathrm{out}}\) is the output projection matrix. The outputs from different kernel sizes are combined using learnable normalized weights:
{
\abovedisplayskip=3pt plus 0pt minus 0pt
\belowdisplayskip=3pt plus 0pt minus 0pt
\abovedisplayshortskip=1pt plus 1pt minus 1pt
\belowdisplayshortskip=1pt plus 1pt minus 1pt
\begin{align}
\mathrm{Conv}(Z) &= \sum_{i=1}^{n}
\frac{\exp(\beta_i)}{\sum_{j=1}^{n}\exp(\beta_j)}
\mathrm{Conv}_{q_i}(Z)
\label{eq22}
\end{align}
}
where \(q_i\) denotes the \(i\)-th convolution kernel size, \(\beta_i\) is its learnable weight, and \(n\) is the number of convolution kernels.

Third, the position-wise feed-forward network applies the feed-forward transformation to each residue representation individually.
{
\abovedisplayskip=3pt plus 0pt minus 0pt
\belowdisplayskip=3pt plus 0pt minus 0pt
\abovedisplayshortskip=1pt plus 1pt minus 1pt
\belowdisplayshortskip=1pt plus 1pt minus 1pt
\begin{align}
\mathrm{Pointwise}(Z) &= \mathrm{FFN}(Z)
\label{eq23}
\end{align}
}
Finally, the outputs of the three branches are summed to obtain the fused representation:
{
\abovedisplayskip=3pt plus 0pt minus 0pt
\belowdisplayskip=3pt plus 0pt minus 0pt
\abovedisplayshortskip=1pt plus 1pt minus 1pt
\belowdisplayshortskip=1pt plus 1pt minus 1pt
\begin{align}
Z_{\mathrm{muse}} 
= \mathrm{Attention}(Z) + \mathrm{Conv}(Z) + \mathrm{Pointwise}(Z)
\label{eq:zmuse}
\end{align}
}
The fused representation \(Z_{\mathrm{muse}}\) is then fed into the final classification head for residue-level PPIS prediction.

\subsection{Focal loss}\label{subsec:Focal loss}

Focal loss was originally proposed to address severe class imbalance by down-weighting well-classified examples and focusing training on hard samples~\cite{lin2017focal}. Because the number of non-interacting residues substantially exceeds that of interacting residues, PPIS prediction is a highly imbalanced binary classification task. In this setting, Focal loss mitigates the effect of class imbalance by assigning larger weights to the minority class while further down-weighting easy examples through the modulating factor. For binary classification, the Focal loss is defined as:
{
\abovedisplayskip=3pt plus 0pt minus 0pt
\belowdisplayskip=3pt plus 0pt minus 0pt
\abovedisplayshortskip=1pt plus 1pt minus 1pt
\belowdisplayshortskip=1pt plus 1pt minus 1pt
\begin{align}
\text{FL}(p_t) &= - \Phi_t (1 - p_t)^\gamma \log(p_t) \label{eq26}
\end{align}
}
As shown in Equation~\ref{eq26}, \(p_t\) denotes the predicted probability of the target class. Specifically, \(p_t=p\) for positive samples and \(p_t=1-p\) for negative samples. Here, \(\Phi_t\) denotes the class-dependent balancing factor, and \(\gamma\) is the focusing parameter. Following previous studies~\cite{fu2024agf,han2025mgma} and empirical validation, non-interacting residues were assigned a balancing factor of 0.25, whereas interacting residues were assigned a balancing factor of 0.75, and \(\gamma\) was set to 2.0.

\section{Results and discussion}\label{sec:Results and discussion}

This section evaluates the predictive performance of SGAP-PPIS on multiple independent test sets. SGAP-PPIS was compared with representative PPIS prediction methods. 

\subsection{Evaluation metrics}\label{subsec:Evaluation metrics}

Performance was evaluated using accuracy (ACC), precision, recall, F1-score (F1), Matthews correlation coefficient (MCC), area under the receiver operating characteristic curve (AUROC), and area under the precision-recall curve (AUPRC). In protein data, the number of non-interacting residues far exceeds that of interacting residues; therefore, more robust metrics for this class imbalance, such as F1, AUROC, AUPRC, and MCC, are more meaningful for evaluation. More detailed definitions and calculation formulas of the evaluation metrics are provided in the Supplementary Material.

\subsection{Experiment details}\label{subsec:Experiment details}
 
In this study, the model was implemented in PyTorch. The model was optimized using the Adam optimizer for up to 50 epochs. During model development, a learning rate decay strategy was adopted. If the validation AUPRC did not improve for 5 consecutive epochs, the learning rate was reduced by 70\% of its previous value, with a lower bound set to \(1 \times 10^{-6}\).
\begin{table}[h]
\centering
\caption{Hyperparameters for SGAP-PPIS}
\begin{tabular}{lccc}
\hline
Hyperparameter  & Value & Hyperparameter & Value \\
\hline
Distance threshold  &14 \AA  & Epoch & 50 \\
APPNP\_Layer  & 5  & EGNN\_Layer & 5 \\
\(d_a\)  & 128  & \(d_g\) & 256 \\
Learning Rate & 0.0001  & Dropout& 0.3 \\
\hline
\end{tabular}

\label{tab:hyperparameters}
\end{table}

\begin{table*}[htb]
\centering
\caption{Performance Comparison with Other Methods on Test\_60.}
\resizebox{0.98\linewidth}{!}{
\begin{tabular}{lccccccc}
\hline
Method & ACC & Precision & Recall & F1 & MCC & AUROC & AUPRC \\
\hline
PSIVER & 0.561 & 0.188 & 0.534 & 0.278 & 0.074 & 0.573 & 0.190 \\
ProNA2020 & 0.738 & 0.275 & 0.402 & 0.326 & 0.176 & N/A & N/A \\
SCRIBER & 0.667 & 0.253 & 0.568 & 0.350 & 0.193 & 0.665 & 0.278 \\
DLPred & 0.682 & 0.264 & 0.565 & 0.360 & 0.208 & 0.677 & 0.294 \\
DELPHI& 0.697 & 0.276 & 0.568 & 0.372 & 0.225 & 0.699 & 0.319 \\
DeepPPISP & 0.657 & 0.243 & 0.539 & 0.335 & 0.167 & 0.653 & 0.276 \\
SPPIDER & 0.752 & 0.331 & 0.557 & 0.415 & 0.285 & 0.755 & 0.373 \\
MaSIF-site & 0.780 & 0.370 & 0.561 & 0.446 & 0.326 & 0.775 & 0.439 \\
GraphPPIS & 0.776 & 0.368 & 0.584 & 0.451 & 0.333 & 0.786 & 0.429 \\
DeepProSite & 0.842 & 0.501 & 0.443 & 0.470 & 0.379 & 0.813 & 0.490 \\
AGAT-PPIS & 0.856 & 0.539 & 0.603 & 0.569 & 0.484 & 0.867 & 0.574 \\
AGF-PPIS & 0.860 & 0.551 & 0.620 & 0.584 & 0.501 & 0.870 & 0.599 \\
GHGPR-PPIS & 0.860 & 0.551 & 0.620 & 0.583 & 0.501 & 0.869 & 0.596 \\
ASCE-PPIS  & $0.856 \pm 0.008$&$0.539 \pm 0.022$ & $0.641 \pm 0.003$ & $0.585 \pm 0.012$ & $0.502 \pm 0.015$& $0.876 \pm 0.001$ & $0.597 \pm 0.014$  \\
MGMA-PPIS  & $0.872\pm 0.003$& $0.586 \pm 0.008$& $0.640 \pm 0.007$& $0.612 \pm 0.006$& $0.536 \pm 0.008$& $0.889 \pm 0.002$& $0.647 \pm 0.009$\\
ComGAT-PPIS & $0.822 \pm 0.015$ & $0.457 \pm 0.030$ & $0.636 \pm 0.021$ & $0.531 \pm 0.015$ & $0.435 \pm 0.019$ & $0.847 \pm 0.009$ & $0.527 \pm 0.013$  \\
\textbf{SGAP-PPIS} & $\boldsymbol{0.879} \pm 0.010$& $\boldsymbol{0.612} \pm 0.039$&$\boldsymbol{ 0.645} \pm 0.028$ & $\boldsymbol{0.627} \pm 0.012$& $\boldsymbol{0.555} \pm 0.016$& $\boldsymbol{0.894} \pm 0.005$& $\boldsymbol{0.670} \pm 0.014$\\
\hline
\end{tabular}
}
\begin{tablenotes}
\item The results of ASCE-PPIS, MGMA-PPIS, ComGAT-PPIS, and SGAP-PPIS were obtained under the unified evaluation protocol used in this study, whereas the results of the remaining methods were taken from the corresponding publications. N/A indicates that there is no available data.
\end{tablenotes}
\label{tab:Test60}
\end{table*}

\begin{table*}
\caption{Performance comparison of different methods on the Test\_315-28 and UBtest\_31-6 datasets.}
\centering
\resizebox{0.98\linewidth}{!}{
\begin{tabular*}{\textwidth}{@{\extracolsep{\fill}}lcccc@{\extracolsep{\fill}}}
\hline
& \multicolumn{2}{@{}c@{}}{Test\_315-28} & \multicolumn{2}{@{}c@{}}{UBtest\_31-6} \\
\cline{2-3}\cline{4-5}
Method &AUPRC & MCC & AUPRC & MCC \\
\hline
ASCE-PPIS & $0.598 \pm 0.009$ & $0.509 \pm 0.016$ & $0.350 \pm 0.027$ & $0.328 \pm 0.030$ \\
MGMA-PPIS         & $0.584 \pm 0.040$ & $0.499 \pm 0.033$ & $0.381 \pm 0.011$ & $0.367 \pm 0.012$ \\
ComGAT-PPIS     & $0.494 \pm 0.032$ & $0.420 \pm 0.028$  & $0.305 \pm 0.024$ & $0.286 \pm 0.033$ \\
    \textbf{SGAP-PPIS }  & $\boldsymbol{0.625} \pm 0.007$ & $\boldsymbol{0.529} \pm 0.006$ & $\boldsymbol{0.408} \pm 0.030$ & $\boldsymbol{0.368} \pm 0.024$ \\
\hline
\end{tabular*}
}
\begin{tablenotes}
\item 
\end{tablenotes}
\label{tab:other test sets}
\end{table*}

In addition, focal loss was used as the training objective. For each random seed, 5-fold cross-validation was first performed on the processed Train\_335-1 dataset, with four folds used for training and one fold used for validation. In each fold, the model was trained for up to 50 epochs, and the checkpoint with the highest validation AUPRC was selected. The average of the fold-wise best epochs was then used as a reference during full-model training on the complete training set under the same hyperparameter configuration. The resulting full-model checkpoints were subsequently evaluated on the independent test sets. All experiments were conducted on a Tesla V100 GPU. The final hyperparameter configuration is shown in Table~\ref{tab:hyperparameters}, and more detailed experimental settings are provided in the Supplementary Material.

\subsection{Performance comparison with other methods}\label{subsec:Performance comparison with other methods}

This study compares SGAP-PPIS on Test\_60 with representative sequence-based and structure-based PPIS prediction methods, including PSIVER~\cite{murakami2010applying}, ProNA2020~\cite{qiu2020prona2020}, SCRIBER~\cite{zhang2019scriber}, DLPred~\cite{zhang2019sequence}, DELPHI~\cite{li2021delphi}, DeepPPISP~\cite{zeng2020protein}, SPPIDER~\cite{porollo2007prediction}, MaSIF-site~\cite{gainza2020deciphering}, GraphPPIS~\cite{yuan2022structure}, DeepProSite~\cite{fang2023deepprosite}, AGAT-PPIS~\cite{zhou2023agat}, AGF-PPIS~\cite{fu2024agf}, GHGPR-PPIS~\cite{zeng2024ghgpr}, ASCE-PPIS~\cite{shen2025asce}, MGMA-PPIS~\cite{han2025mgma} and ComGAT-PPIS~\cite{zhang2025comgat}. 
For a controlled comparison, ASCE-PPIS was evaluated using the same 62-dimensional feature setting as SGAP-PPIS. ASCE-PPIS, MGMA-PPIS, ComGAT-PPIS, and SGAP-PPIS were evaluated under the same protocol with three random seeds, and the reported results are presented as $mean \pm std$ deviation over the three runs.

Table~\ref{tab:Test60} presents a comparison against both graph-based baselines and previously reported representative methods. Non-GNN models cannot fully exploit protein molecular structure information, whereas existing Graph-based deep learning methods often rely on fixed propagation mechanisms. Such fixed propagation schemes may not satisfy the requirement that different residues need different balances between neighborhood diffusion and feature preservation, thereby limiting their ability to learn residue-specific propagation patterns. Compared with MGMA-PPIS, SGAP-PPIS improves F1 by 0.015, MCC by 0.019, and AUPRC by 0.023 on the Test\_60. To compare its performance on the independent test sets Test\_315-28 and UBtest\_31-6 with these methods, the experimental results are shown in Table~\ref{tab:other test sets}. Compared with MGMA-PPIS, SGAP-PPIS improves AUPRC by 0.041 and MCC by 0.030 on Test\_315-28, and improves AUPRC by 0.027 on UBtest\_31-6.

\section{Ablation experiment}\label{sec:Ablation experiment}

To systematically evaluate the contribution of each key component in SGAP-PPIS, six ablation experiments were conducted across the same training, validation, and testing protocols as the full model. These ablations examine the effects of geometry-conditioned adaptive propagation, propagation-step and geometric-scale alignment, multi-step APPNP state concatenation, the APPNP branch, the EGNN branch, and the MUSE fusion module. For the ablation study, all variants were trained and evaluated using the same data splits, hyperparameter settings, and random seed. Specifically, Table~\ref{tab:ablation_SGAP} reports the results obtained with the same random seed.

\subsection{The Impact of Geometry-Conditioned Adaptive Propagation}\label{subsec:The Impact of Geometry-Conditioned Adaptive Propagation}

A fixed-coefficient variant, denoted SGAP\_Fixed\_\(\alpha\), was constructed to evaluate the role of geometry-conditioned adaptive propagation.
In this setting, all residues and propagation steps share the same coefficient, allowing us to examine whether the performance gain comes from geometry-conditioned regulation rather than the APPNP backbone alone. 

\begin{figure}[h]
    \centering
    \includegraphics[width=0.80\linewidth]{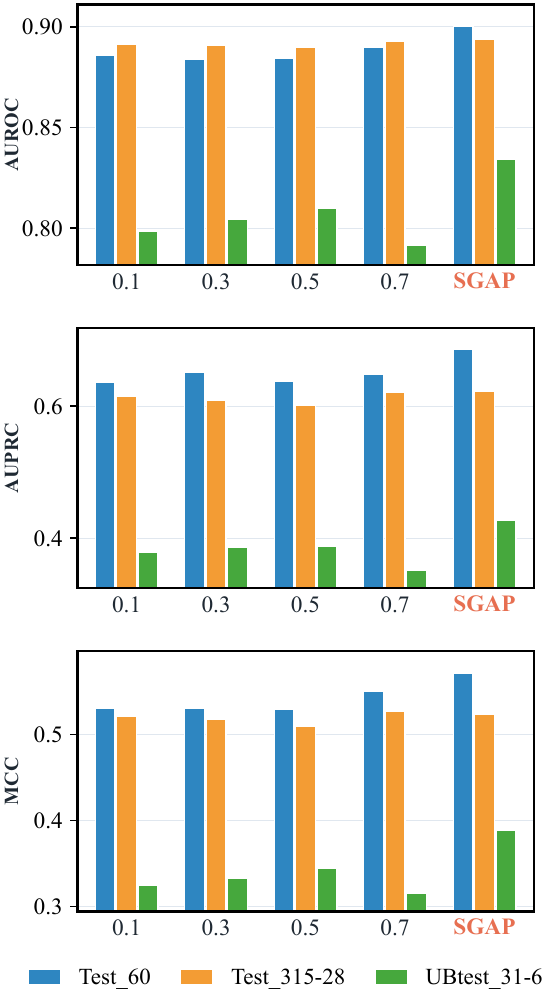}
    \caption{Effect of different fixed residue-wise propagation coefficients on AUROC, AUPRC, and MCC across three test sets, with SGAP-PPIS shown for comparison.}
    \label{fig:fixed_alpha}
\end{figure}

To avoid relying on a single hyperparameter setting, a parameter sweep over $\mathrm{APPNP}\_\alpha\in \{0.1, 0.3, 0.5, 0.7\}$ was conducted. This range spans regimes from stronger neighborhood diffusion to greater feature preservation.
As shown in Fig.~\ref{fig:fixed_alpha}, the SGAP-PPIS outperforms fixed-coefficient variants on the three independent test sets. The optimal static coefficient varies across datasets, suggesting that a globally fixed residue-wise propagation coefficient cannot adequately adapt to diverse protein structures.

\subsection{The Impact of Propagation-Step and Geometric-Scale Alignment}\label{subsec:The Impact of Propagation-Step and Geometric-Scale Alignment}
To investigate the role of the propagation-step and geometric-scale alignment, a w/o ScaleAlign variant was constructed. In this ablation setting, the residue-wise propagation coefficient $\mathrm{APPNP}\_\alpha$ is no longer generated from the geometric context aligned with the current propagation step. Instead, a global multi-scale geometric summary is uniformly used as the shared guidance signal for all propagation steps. 
\begin{table*}[t]
\caption{Results of the ablation experiments for SGAP-PPIS on the Test\_60, Test\_315-28, and UBtest\_31-6 datasets.}
\label{tab:ablation_SGAP}
\centering
\setlength{\tabcolsep}{8pt}
\begin{tabular}{lccccccccc}
\hline
& \multicolumn{3}{c}{Test\_60} & \multicolumn{3}{c}{Test\_315-28} & \multicolumn{3}{c}{UBtest\_31-6} \\
\cmidrule(lr){2-4}\cmidrule(lr){5-7}\cmidrule(lr){8-10}
Method & AUROC & AUPRC & MCC & AUROC & AUPRC & MCC & AUROC & AUPRC & MCC \\
\hline
w/o ScaleAlign        & 0.888 & 0.652 & 0.538 & 0.891 & 0.617 & 0.519 & 0.796 & 0.375 & 0.327 \\
w/o Multi-Step Concat & 0.883 & 0.635 & 0.525 & 0.889 & 0.614 & 0.516 & 0.808 & 0.379 & 0.351 \\
w/o APPNP             & 0.885 & 0.643 & 0.525 & 0.888 & 0.616 & 0.516 & 0.795 & 0.360 & 0.341 \\
w/o EGNN              & 0.824 & 0.494 & 0.420 & 0.829 & 0.491 & 0.399 & 0.764 & 0.322 & 0.295 \\
w/o MUSE              & 0.885 & 0.649 & 0.528 & 0.892 & 0.608 & 0.521 & 0.804 & 0.384 & 0.333 \\
\textbf{SGAP-PPIS }         & \textbf{0.900} & \textbf{0.686} & \textbf{0.571} & \textbf{0.894} & \textbf{0.622} & \textbf{0.523} & \textbf{0.834} & \textbf{0.428} & \textbf{0.389} \\
\hline

\end{tabular}
\end{table*}

The detailed results are reported in Table~\ref{tab:ablation_SGAP}. Removing ScaleAlign reduced Test\_60 AUPRC from 0.686 to 0.652 and MCC from 0.571 to 0.538, with similar decreases on UBtest\_31-6. 
Therefore, the results of w/o ScaleAlign further suggest that the full model does not simply apply a single geometric feature uniformly to all propagation steps. 
Instead, effective propagation control in PPIS prediction depends on geometric-scale alignment with propagation-step, rather than applying a uniform geometric signal throughout the entire APPNP process.

\subsection{The Impact of Multi-step APPNP State Concat}\label{subsec:The Impact of Multi-step APPNP State Concat}

To evaluate the contribution of multi-step APPNP state concatenation, a w/o Multi-Step Concat variant was constructed, which uses only the final APPNP output for subsequent fusion. In the full SGAP-PPIS, the APPNP branch preserves intermediate propagation states and concatenates them for subsequent feature fusion. This setting is designed to examine whether the model's predictive performance depends on hierarchical topological representations generated by multi-step propagation states, rather than solely on the final APPNP output.

The experimental results are summarized in Table~\ref{tab:ablation_SGAP}. Compared with the full model, w/o Multi-Step Concat reduced AUPRC from 0.686 to 0.635 on Test\_60 and from 0.428 to 0.379 on UBtest\_31-6. These results suggest that relying solely on the final APPNP output results in the loss of local and intermediate-range topological context captured at earlier propagation stages, thereby weakening the APPNP branch's ability to model topological context across different propagation depths. 

\subsection{Model Structure Ablation Experiment}\label{subsec:Model Structure Ablation Experiment}

Further model-level ablations were conducted by removing the APPNP branch, the EGNN branch, or the MUSE fusion module.

The experimental results are presented in Table~\ref{tab:ablation_SGAP}. Compared with the full SGAP-PPIS, both w/o APPNP and w/o EGNN exhibit performance degradation to different extents. This suggests that the APPNP and EGNN branches provide complementary information. Among these variants, w/o EGNN shows the largest performance drop, indicating that the EGNN branch not only provides geometric information but also provides the geometric context used to generate residue-wise propagation coefficients. The results of w/o MUSE show that even when the dual-branch structure is retained, omitting the MUSE module for further cross-branch feature fusion still degrades model performance. This finding indicates that the performance advantage of SGAP-PPIS is not only derived from the joint introduction of APPNP-based topological information and EGNN-based geometric information, but also depends on an effective cross-branch fusion mechanism.
\begin{figure*}[ht]
\centering
\includegraphics[width=0.9\linewidth]{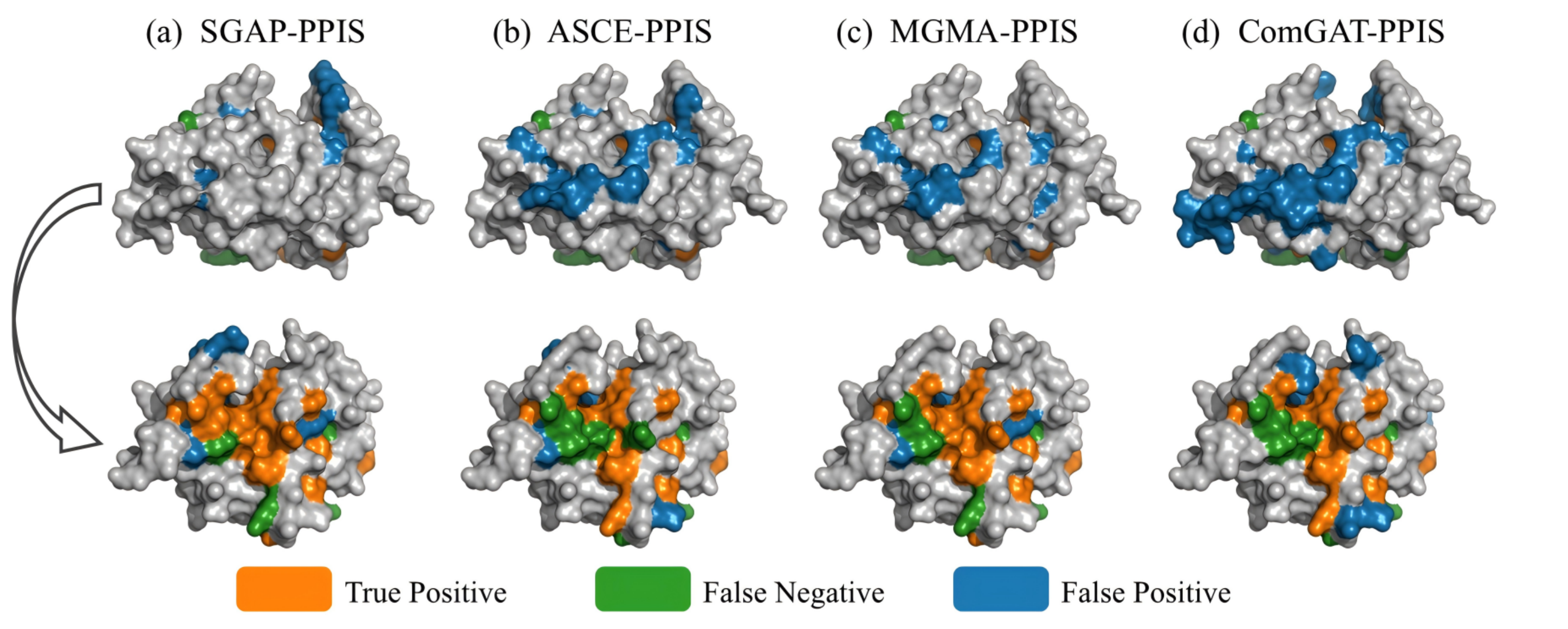}
\caption{Representative prediction results for protein 4BVX chain B. 
(a) Interaction sites predicted by SGAP-PPIS; 
(b) interaction sites predicted by ASCE-PPIS; 
(c) interaction sites predicted by MGMA-PPIS; 
(d) interaction sites predicted by ComGAT-PPIS.}
\label{fig:4BVXB}
\figalttext[Representative prediction results for protein 4BVX chain B.]%
{The figure compares residue-level interaction site predictions for protein 4BVX chain B using SGAP-PPIS, ASCE-PPIS, MGMA-PPIS, and ComGAT-PPIS.}%
\end{figure*}
Taken together, the performance gains of SGAP-PPIS stem from the coordinated contributions of geometry-conditioned adaptive propagation, scale-aligned geometric guidance, multi-step topological representation, and MUSE-based feature fusion.
\section{Case study}\label{sec:Case study}

A representative case study was conducted to qualitatively illustrate the prediction performance of SGAP-PPIS on an individual protein chain.
This protein is an important cofactor of the polyaminoacyl-tRNA synthetase complex, participates in complex assembly, and is implicated in regulating the DNA damage response. 

Table~\ref{tab:4BVXA} presents the site-level prediction results of SGAP-PPIS, ASCE-PPIS, MGMA-PPIS, and ComGAT-PPIS. SGAP-PPIS identifies more true-positive and true-negative sites while producing fewer false positives and false negatives.

As shown in Fig.~\ref{fig:4BVXB}, the prediction results were further visualized using PyMOL. From the colored regions, SGAP-PPIS produces fewer false-positive predictions in the interface regions, indicating that geometry-conditioned adaptive propagation may help reduce false-positive predictions near the interface.
\begin{table}[ht]
\centering
\caption{Site-level prediction results of different methods on EEF1E1/AIMP3 (PDB ID: 4BVX, chain B).}
\label{tab:4BVXA}
\begin{tabular*}{\linewidth}{@{\extracolsep{\fill}}lcccc}
\hline
Model & TP & TN & FP & FN \\
\hline
\textbf{SGAP-PPIS} & \textbf{26} & \textbf{126} & \textbf{10} & \textbf{8} \\
ASCE-PPIS & 20 & 115 & 21 & 14 \\
MGMA-PPIS & 23 & 117 & 19 & 11 \\
ComGAT-PPIS & 21 & 104 & 32 & 13 \\
\hline
\end{tabular*}
\end{table}

\section{Conclusion}\label{sec:Conclusion}

This paper presents SGAP-PPIS, a structure-guided adaptive propagation model for PPIS prediction. By deriving residue-wise propagation coefficients from multi-scale geometric states, SGAP-PPIS allows local geometric context to regulate the balance between feature preservation and neighborhood diffusion during graph learning.

Experimental evaluation across three benchmark test sets shows that SGAP-PPIS performs competitively on Test\_60 and achieves the highest AUPRC on both Test\_315-28 and UBtest\_31-6 among the re-evaluated methods. Ablation analyses further suggest that these gains are primarily associated with geometry-conditioned adaptive propagation, supported by scale-aligned geometric guidance, multi-step topological representation, and MUSE-based feature fusion.

While validated on benchmark datasets, applying SGAP-PPIS to predict interactions in dynamic, multi-component protein complexes or on membrane proteins remains an open challenge. Future work will explore integrating temporal dynamics or lipid bilayer context into the adaptive propagation model.

\section*{Acknowledgments}

This work was supported by the National Major Scientific Instrument and Equipment Development Project of National Natural Science Foundation of China (No. 62427811), the National Natural Science Foundation of China (No. 62272115).

\bibliographystyle{IEEEtran}
\bibliography{reference}

\end{document}